\documentclass[12pt]{article}

\usepackage{amsmath}
\usepackage{tikz-cd}
\usepackage{amssymb}
\usepackage{graphicx}
\usepackage{hyperref}
\usepackage{geometry}
\usepackage{mathtools}
\usepackage{titlesec}
\usepackage{titling}
\usepackage[backend=bibtex,style=numeric,sorting=none,defernumbers=true]{biblatex}
\usepackage{etoolbox}   

\titleformat{\section}      {\normalfont\Large\bfseries\centering}{\thesection}{1em}{}
\titleformat{\subsection}   {\normalfont\large\bfseries\centering}{\thesubsection}{1em}{}
\titleformat{\subsubsection}{\normalfont\normalsize\bfseries\centering}{\thesubsubsection}{1em}{}

\titleclass{\subsubsubsection}{straight}[\subsubsection]
\newcounter{subsubsubsection}[subsubsection]
\renewcommand\thesubsubsubsection{\thesubsubsection.\arabic{subsubsubsection}}
\titleformat{\subsubsubsection}
  {\normalfont\normalsize\bfseries\centering}{\thesubsubsubsection}{1em}{}
\titlespacing*{\subsubsubsection}{0pt}{1.5ex plus 0.5ex minus .2ex}{1em}

\pretitle{%
  \begin{center}
    \vspace*{-0.42cm}
    \rule{\linewidth}{3pt}\\[\smallskipamount]
    \bfseries\LARGE
    \vspace{0.22cm}
}
\posttitle{%
    \\[\smallskipamount]
    \rule{\linewidth}{1.6pt}
  \end{center}
  \vspace{0.12cm}
}

\preauthor{\begin{center}\large}
\postauthor{\end{center}}

\predate{\begin{center}\large}
\postdate{\end{center}}

\makeatletter
\g@addto@macro\maketitle{\thispagestyle{empty}}
\makeatother

\title{Higher Gauge Flow Models}

\author{%
  \begin{minipage}[t]{0.46\textwidth}\centering
    Alexander Strunk\thanks{Corresponding author: \href{mailto:astrunk.research@evercot.ai}{astrunk.research@evercot.ai}}\\
    Evercot AI\\
  \end{minipage}%
  \hspace{1em}
  \begin{minipage}[t]{0.46\textwidth}\centering
    Roland Assam\\
    Evercot AI\\
  \end{minipage}%
\vspace{0.32cm}
}

\date{18 July 2025}

\begin{document}

\maketitle
\thispagestyle{empty} 
\vspace{-0.60cm}
\begin{abstract}
\noindent 
This paper introduces Higher Gauge Flow Models, a novel class of 
Generative Flow Models. Building upon ordinary Gauge Flow Models \cite{GaugeFlowModels},  these Higher Gauge Flow Models leverage an 
L$_{\infty}$-algebra, effectively extending the Lie Algebra. This 
expansion allows for the integration of the higher geometry and higher 
symmetries associated with higher groups into the framework of Generative 
Flow Models.  Experimental evaluation on a Gaussian Mixture Model dataset 
revealed substantial performance improvements compared to traditional Flow 
Models.\\

\end{abstract}

\newpage 
\setcounter{page}{1} 

\section{Introduction}
A Higher Gauge Flow Model is a novel approach to generative modeling,  where the dynamics is governed by the following neural Ordinary Differential Equation (ODE):
\begin{align*}
\hat{\nabla}_{dt} x(t) := v_{\theta}(x(t),  t) - \alpha(t) \Pi_{M, \hat{W}} \Bigl( \mathcal{A}_{\mu}(x(t),  t) \bigl[  \hat{v}(x(t), t) \bigr]  d^{\mu}(x(t), t)  \Bigr)
\end{align*} 
where $\hat{v}(x(t), t)$ has the structure of a graded vector.  The Higher Gauge Field acts on this graded vector as follows:
\begin{align*}
  \mathcal{A}_{\mu}(x(t),  t) \bigl[  \hat{v}(x(t), t) \bigr]  d^{\mu}(x(t), t) :=  \sum_m \mathcal{A}^{a}_{\mu}(x(t),  t) d^{\mu}(x(t), t) b_{m} ( e_{a},  \hat{v}(x(t), t),  \hdots ,  \hat{v}(x(t), t))
\end{align*} 

\noindent Here,  the index $m$ represents the number of inputs for the higher brackets $b_{m}$ within the L$_{\infty}$-algebra,  a sophisticated algebraic structure providing a framework for representing and manipulating higher-order symmetries and invariants.  The Higher Gauge Field,  denoted as $\mathcal{A}_{\mu}(x(t),  t)$,  is valued within the graded vector space of the L$_{\infty}$-algebra.  This novel incorporation of an L$_{\infty}$-algebra enables the generative model to integrate richer mathematical structures.  Specifically,  this framework enables the exploration of interesting model architectures and potentially introduces higher symmetries into the domain of deep learning.

\section{Mathematical Background}
This section introduces the mathematical foundations of Higher Gauge Flow 
Models,  which represent a further development of Gauge Flow Models \cite{GaugeFlowModels}.  The fundamental mathematical innovation behind Higher Gauge Flow Models lies in the application of L$_{\infty}$-algebras – complex mathematical structures.  To fully appreciate the potential for future 
research and development in this area, a basic understanding of Higher Differential Geometry and Higher Category Theory is recommended.   For introductory materials or a refresher, the following resources might be helpful:
\begin{itemize}
\item \textbf{Higher Differential Geometry}: \cite{HigherGeometryPhysics, HigherGeometricStructuresManifolds,DifferentiableGerbes}
\item \textbf{Higher Gauge Theory}: \cite{PrincipalInfBundles, LInfAlgebraHPT, HigherGaugeTheory}
\item  \textbf{Higher Category Theory}: \cite{HigherToposTheoryPhysics,  InfinityTopos,  HigherStructuresMTheory}
\end{itemize}

\subsection{Graded Vector Space}
A graded vector space is a vector space equipped with a decomposition into a direct sum of subspaces indexed by a set,  usually the integers.  Formally,  a graded vector space $\hat{V}$ is defined as:
\[
\hat{V} = \bigoplus_{i \in I} V_{i}
\]
where I is an index set (usually an abelian group such as $\mathbb{Z}$) and each $V_i $ is a subspace of V.  A vector that lies entirely in a single $V_{i}$ (i.e.  in just one degree) is called a homogeneous element of degree $i$.  Graded vector spaces are fundamental throughout algebra and topology, appearing in chain complexes, polynomial and exterior algebras, graded Lie (and super-) algebras, and in the study of homology, cohomology, and generating functions.\\
When the set $I$ is equipped with an abelian group operation $+ : I \times I \rightarrow I$,  this additional structure enables the definition of the dual graded vector space $\hat{V}^{\ast}$ of $\hat{V}$ precisely as follows:
\[
\hat{V}^{\ast} := \bigoplus_{i \in I } (\hat{V}^{\ast})_{i}, \quad \text{where} \quad (\hat{V}^{\ast})_{i} := (V_{-i})^{\ast}
\]
where $-i$ represents the inverse with respect to the group operation $+$.\\
A Tensor product $\hat{\otimes}$ of two graded vector spaces $\hat{V}, \hat{W}$ graded by the same abelian group $I$,  with operation $+$,  is defined as:
\[
\hat{V} \hat{\otimes} \hat{W} = \bigoplus_{m \in I}  (\hat{V} \hat{\otimes} \hat{W})_{m}, \quad \text{where} \quad  (\hat{V} \hat{\otimes} \hat{W})_{m} := \bigoplus_{i + j = m} (V_{i} \otimes W_{j})
\]
An additional structure that can be introduced is a braiding,  which defines a mapping for swapping vector spaces.  Assuming a parity map or a typical $I = \mathbb{Z}$ grading, a braiding $\mathcal{B}$ is defined as:
\begin{align*}
\mathcal{B}: &\hat{V} \hat{\otimes} \hat{W}  \rightarrow \hat{W} \hat{\otimes} \hat{V} \\
& v \hat{\otimes} w \mapsto  (-1)^{f(|v|,  |w|)} w \hat{\otimes} v
\end{align*}
where $f: \mathbb{Z} \times \mathbb{Z} \rightarrow \mathbb{Z}$ is a function depending on the degrees $|v|, |w| \in \mathbb{Z}$  of the individual homogeneous elements $v, w$ of the vector spaces $V_{|v|}, W_{|w|}$.  In this paper,  where we heavily rely on $I=\mathbb{Z}$, $f$ is defined as:
 \[
f(|v|,  |w|) := |v| \cdot |w|
\]
It is crucial to note that linear maps (morphisms) $\hat{F}: \hat{V} \rightarrow \hat{W}$ between two graded vector spaces also possess a graded vector space structure,  as $\hat{F} \in \hat{W} \hat{\otimes} \hat{V}^{\ast}$.  

\subsubsection{L$_{\infty}$-algebra}

An L$_{\infty}$-algebra represents a generalization of a Lie algebra,  designed to encode higher geometric and algebraic structures.  Instead of requiring the strict Jacobi identity,  it replaces that single constraint with an infinite,  coherent hierarchy of identities linked by higher homotopies.  This added flexibility is indispensable in modern settings-deformation theory, topological field theory,  derived geometry—where rigid equalities are often too brittle to capture the phenomena of interest.
Mathematically,  an L$_{\infty}$-algebra $\hat{L}$ is defined as a $\mathbb{Z}$-graded vector space:
\[
\hat{L}  = \bigoplus_{i \in \mathbb{Z}}  L_{i}
\]
Its structure is determined by higher brackets $b_{m}$ with $m \geq 1$. These brackets are multilinear, graded skew-symmetric maps:
\begin{align*}
b_{m}: & \hat{L} \times \dots \times \hat{L} \rightarrow \hat{L}\\
& (l_1,  \dots ,  l_m) \mapsto b_m(l_1, \dots, l_m)
\end{align*}
The degree of each higher bracket $b_{m}$ is $m-2$.  
The structure $\hat{L}$ is determined by the following:
\begin{itemize}
\item \textbf{Graded Skew-Symmetry:} The higher brackets $b_m$ satisfy the following relation:
\[
 b_m(l_1, \dots, l_m) = \Xi(\sigma)  b_m(l_{\sigma(1)}, \dots, l_{\sigma(m)})
\]
where the elements $l_1,  \hdots, l_m \in L_{|l_1|}, \hdots , L_{|l_m|}$ are homogeneously graded. \\
The sign $\Xi(\sigma) \in \{1, -1\}$ is determined by the graded Koszul sign rule:
\[
l_1 \wedge \hdots \wedge l_m = \Xi(\sigma) l_{\sigma(1)} \wedge \hdots \wedge l_{\sigma(m)}
\]
Here, $\sigma \in \mathcal{S}_m$ represents a permutation of $m$ elements, and the wedge product is induced by the graded tensor product $\hat{\otimes}$ for the two homogeneous elements $l_1, l_2 \in L_{|l_1|}, L_{|l_2|}$:
\[
l_1 \wedge l_2 = l_1 \hat{\otimes} l_2 - (-1)^{|l_1| |l_2|} l_2 \hat{\otimes} l_1
\]
\item \textbf{Higher Jacobi Identity:} For homogeneous elements $l_1,  \hdots, l_k \in L_{|l_1|}, \hdots , L_{|l_k|}$, the higher Jacobi identity is given by:
\[
\sum_{m+n = k +1} \sum_{\sigma \in \mathcal{S}_k|_c} \Xi(\sigma) (-1)^{m (n-1)} b_{n}(b_{m}(l_{\sigma(1)}, \hdots, l_{\sigma(m)}), l_{\sigma(m+1)},  \hdots , l_{\sigma(k)}) = 0
\]
Here, the permutation $\sigma \in \mathcal{S}_k$ is constrained by the condition $c$ for $1 \leq m \leq k$:
\begin{gather*}
\sigma(1) < \hdots < \sigma(m) \\
\sigma(m +1) < \hdots < \sigma(k)
\end{gather*}
\end{itemize}
\noindent There are several equivalent formulations of the mathematical structure of an L$_{\infty}$-algebra. For more information refer to \cite{LInfinityAlgebras,  StrongHomotopyAlgebras}.

\newpage
\section{Higher Gauge Flow Model}
A Higher Gauge Flow Model is defined on a graded vector bundle $\hat{A}$:
\begin{center}
\begin{tikzcd}[column sep=3.5em,row sep=3.5em]
& {\scalebox{1.5}{$\substack{\hat{A}\\=\\E\times_M F}$}}
    \arrow[dl, swap, "p_E"]
    \arrow[dr, "p_F"]
& \\[-0.2em]
E
    \arrow[dr, swap, "\pi_E"]
& 
& F
    \arrow[dl, "\pi_F"] \\[0.2em]
& M &
\end{tikzcd}
\end{center}
with the following:
\begin{itemize}
\item The base manifold M.
\item The graded vector bundle $E$ with the fiber,  a graded vector space $\hat{V}$ with the structure of an L$_{\infty}$-algebra.
\item  The graded vector bundle $F$ with the fiber,  a graded vector space $\hat{W}$ with the structure of an L$_{\infty}$-algebra.
\item The fiber product $E\times_M F \coloneqq \{ (e,f) | \pi_{E}(e) = \pi_{F}(f) \}$.
\end{itemize}
\noindent The dynamics are described by the following differential equation:
\begin{align*}
\hat{\nabla}_{dt} x(t)  = v_{\theta}(x(t),  t) - \alpha(t) \Pi_{M, \hat{W}} \Bigl( \mathcal{A}_{\mu}(x(t),  t) \bigl[  \hat{v}(x(t), t) \bigr]  d^{\mu}(x(t), t)  \Bigr)
\end{align*} 
where the $\hat{v}(x(t), t)$ is now a graded vector.  While the action of the Higher Gauge Field on this is defined as:
\begin{align*}
  \mathcal{A}_{\mu}(x(t),  t) \bigl[  \hat{v}(x(t), t) \bigr]  d^{\mu}(x(t), t) :=  \sum_m \mathcal{A}^{a}_{\mu}(x(t),  t) d^{\mu}(x(t), t) b_{m} ( e_{a},  \hat{v}(x(t), t),  \hdots ,  \hat{v}(x(t), t))
\end{align*} 
Here,  the components are defined as follows:
\begin{itemize}
\item $v_{\theta}(x(t),  t) \in TM$ is a learnable vector field modeled by a Neural Network.
\item $\alpha (t)$ is a time dependent weight,  also modeled by a Neural Network.
\item  $\mathcal{A}_{\mu} (x(t),t) \in \Omega^{1}(M) \hat{\otimes} \hat{V}$ is a higher gauge field,  namely an L$_{\infty}$-algebra valued differential form,  which is also modelled by a neural network.
\item $d^{\mu}(x(t), t) \in TM$ is the direction vector field.
\item $ \hat{v}(x(t), t) $  is a graded vector field valued in $\hat{W}$,  which can be learned by a Neural Network.
\item $\Pi_{M, \hat{W}}: F \rightarrow TM$ is a smooth projection from the vector bundle $F$ to the tangent bundle $TM$ of the base manifold $M$.  
\end{itemize}
Crucially,  the structure of the L$_{\infty}$-algebras $\hat{V}$ and $\hat{W}$ is consistent with the higher brackets $b_m$ operating between these two graded vector spaces.  A potential generalization allows 
multiple vector fields, $\hat{v}(x(t), t)$,  to be learned by Neural Networks.

\noindent The Higher Gauge Flow Model (HGFM) is defined for any differentiable base manifold $M$.  Similar to ordinary Gauge Flow Models \cite{GaugeFlowModels},  training the HGFM requires $M$ to be a Riemannian manifold,  possessing a Riemannian metric $g$. The training procedure for the Higher Gauge Flow Model is identical to that of ordinary Gauge Flow Models.  It employs the Riemannian Flow 
Matching (RFM) framework \cite{RFM}, generalizing Flow Matching (FM) 
\cite{FM}. The training objective,  the HGFM loss,  is defined as:
\[
\mathcal{L}_{\mathrm{HGFM}} =  \mathbb{E}_{\substack{t \sim \mathcal{U}[0,1] \\ x \sim p_t}}  \bigg\lVert \Big[ v_{\theta}(x,  t) - \alpha(t) \Pi_{M, \hat{W}} \Bigl( \mathcal{A}_{\mu}(x,  t) \bigl[  \hat{v}(x, t) \bigr]  d^{\mu}(x, t)  \Bigr) \Big] - u_{t}(x) \bigg\rVert_{g_x}^{2}
\]
where $\|\cdot\|_{g_x}$ is the norm induced by the Riemannian metric $g$ of the base manifold $M$ at point $x$. Both terms lie in the tangent space of the base manifold $M$:
\begin{gather*}
\Big[ v_{\theta}(x,  t) - \alpha(t) \Pi_{M, \hat{W}} \Bigl( \mathcal{A}_{\mu}(x,  t) \bigl[  \hat{v}(x, t) \bigr]  d^{\mu}(x, t)  \Bigr) \Big] \in T_x M\\
 u_{t}(x) \in T_x M 
\end{gather*}
The Higher Gauge Flow Model loss incorporates the following components:
\begin{itemize}
  \item \textbf{Probability-density path}  
      $p_t : M \to \mathbb R_{+}$, $t \in [0,1]$, as defined in~\cite{RFM}.  
      For each~$t$, $p_t$ satisfies
      \[
        \int_{M} p_t(x)\, d\mathrm{vol}_x = 1,
      \]
      where $d\mathrm{vol}_x$ is the Riemannian volume form.
  \item \textbf{Target vector field}  
      $u_t(x)$, obtained from the conditional field $u_t(x \mid x_1)$ by
      \[
        u_t(x) =
        \int_{M}
        u_t\!\bigl(x \mid x_1\bigr)\,
        \frac{p_t\!\bigl(x \mid x_1\bigr)\, q(x_1)}{p_t(x)}
        \, d\mathrm{vol}_{x_1},
      \]
      with $q(x) \coloneqq p_{t=1}(x)$ and $x_1$ denoting the sample at time $t=1$.
\end{itemize}
\noindent In general,  computing the HGFM loss $\mathcal{L}_{\mathrm{HGFM}}$ exactly is computationally intractable.  However,  this loss coincides with the Riemannian Flow Matching (RFM) objective.  To overcome the intractability,~\cite{RFM} introduces the Riemannian Conditional Flow Matching (RCFM) loss—an unbiased, single-sample, Monte Carlo estimator of the RFM objective.  The core idea is to replace the marginal target field $u_t(x)$ with its conditional counterpart $u_t(x \mid x_1)$,  thereby enabling tractable training.  This formulation supports efficient,  simulation-free learning on simple geometries and scalable neural network training on general Riemannian manifolds.  Notably,  this approach can also be applied to train Higher Gauge Flow Models.

\noindent For detailed derivations and implementation specifics,  refer to~\cite{RFM,FM}.
\newpage
\section{Experiments}
This study compares the novel Higher Gauge Flow Models with ordinary Gauge Flow Models \cite{GaugeFlowModels} and ordinary Flow Models,  using a  generated Gaussian Mixture Model (GMM) dataset.  The Higher Gauge Flow Models employ a 2-term L$_{\infty}$-algebra $\hat{L}$,  given by:
\[
\hat{L} = L_{0} \oplus L_{1}
\]
The mathematical structure of the L$_{\infty}$-algebra $\hat{L}$ is as follows:
\begin{itemize}
\item $\mathbf{L_{0}}$: The degree zero vector space, $L_{0} = \mathbb{R}^{N(N-1)/2}$,  is the Lie algebra $\mathfrak{so}(N)$ of the special orthogonal group $SO(N)$.
\item $\mathbf{L_{1}}$: The degree one vector space, $L_{1} =  \mathbb{R}^{N(N-1)/2} \oplus  \mathbb{R} \cdot c$, consists of a second copy of $\mathfrak{so}(N)$ alongside a central scalar $c$.
\item \textbf{Brackets $b_n$}: The higher brackets $b_n$ vanish for all $n > 2$. \\ The non-vanishing brackets are defined as follows:
\begin{itemize}
\item \textbf{Unary bracket $b_1$:} The unary bracket $b_{1}: L_{1} \rightarrow L_{0}$ acts as the identity function $b_1(x) = x$ on the $\mathfrak{so}(N)$ part of $L_1$, and maps the central element to zero: $b_1(c) = 0$.
\item \textbf{Binary bracket $b_2$:}  The bracket $b_{2}: \hat{L} \times \hat{L} \rightarrow \hat{L}$ operates according to the following degrees:
\begin{itemize}
\item \textbf{Degree (0,0):} For two elements $l_1, l_2 \in L_{0}$ the bracket $b_2$ corresponds to  brackets $[-,-]$ of the Lie algebra $\mathfrak{so}(N)$:
\[
b_{2} (l_1,  l_2) := [l_1,l_2]_{\mathfrak{so}(N)}
\]
\item \textbf{Degree (0,1):} For two elements $l_1 \in L_{0}$ and $l_2 \in L_{1}|_{\mathfrak{so}(N)}$ the bracket $b_2$ is the adjoint action:
\[
b_{2} (l_1,  l_2) := [l_1,l_2]_{\mathfrak{so}(N)}
\]
while the bracket between the  $l_1 \in L_{0}$ and an element of the central part $l_2 \in L_{1}|_{c}$ vanishes:
\[
b_{2} (l_1,  l_2) := 0
\]
Thus, the one-dimensional line $ L_{1}|_{c} = \mathbb{R} \cdot c$ is a trivial representation.
\item \textbf{Degree (1,1):} The bracket $b_{2}$ vanishes for all the elements $l_1, l_2 \in L_{1}$ due to the degrees of the elements:
\[
b_{2} (l_1,  l_2) := 0
\]
\end{itemize}
\end{itemize} 
\end{itemize}
The model’s performance is evaluated across different values of $N$ for the 2-term L$_{\infty}$-algebra $\hat{L}$.
\subsection{Model}
The Higher Gauge Flow Models are built upon the following components:
\begin{itemize}
\item \textbf{Graded Vector Bundle:} $\hat{A} = E\times_M F$ where  $E = M \times \hat{L}$ and $F = M \times \hat{L}$ are two trivial graded vector bundles with base manifold $M = \mathbb{R}^{N}$.
\item \textbf{Vector Field:} The vector field $v_{\theta} (x(t), t)$,  defined on the tangent bundle $TM$,  is modeled by a Neural Network.
\item \textbf{Alpha Function:} $\alpha(t)$ is also modeled using a Neural Network.
\item \textbf{Higher Gauge Field:} The Higher Gauge Field $\mathcal{A} \in \Omega^{1} (M) \otimes \hat{L}$, valued in the L$_{\infty}$-algebra $\hat{L}$,  is represented by a Neural Network.
\item \textbf{Directional Vector Field:} $d^{\mu} (x(t), t) $ corresponding to a section of $TM$,  which is modeled by a Neural Network.
\item \textbf{Graded Vector Field:} $\hat{v}(x(t), t) : M \rightarrow M \times \hat{L}$  is modeled using a Neural Network.
\item  \textbf{Projection Map:} The projection map $ \Pi_{M,  \hat{L}} : M \times \hat{L} \rightarrow M \times \mathbb{R}^{N}$ projects the graded vector field from the graded vector space $\hat{L}$ to the tangent space $\mathbb{R}^{N}$ of the base manifold $M$.
\end{itemize}
The Neural Network models used are standard Multi-Layer Perceptrons (MLPs) with the subsequent configurations:
\begin{table}[h!]
\centering
\begin{tabular}{|c|c|c|}
  \hline
  Field & Layer Dimensions & Activation Function \\
  \hline
 $\mathcal{A}$       &   [N + 1,   32/64,  32/64,  $N ( 2 N (N-1) /2 + 1)$ ]    &  SiLU         \\
 $v_{\theta} (x(t), t)$      &   [N + 1,   32/64,  32/64,  N]    &  SiLU        \\
 $d^{\mu} (x(t), t)$ &   [N + 1,   32/64,  32/64,  N]    &  SiLU        \\
 $\hat{v}(x(t), t)|_{L_{0}}$       &   [N + 1,   32/64,  32/64,  $N(N-1)/2$]    &  SiLU         \\
 $\hat{v}(x(t), t)|_{L_{1}}$       &   [N + 1,   32/64,  32/64,  $N(N-1)/2 + 1$]    &  SiLU         \\
 $\Pi_{M,  \hat{L}}|_{L_{0}}$       &   [N + 1,   32/64,  32/64,  $N (N(N-1)/2)$]    &  SiLU         \\
 $\Pi_{M,  \hat{L}}|_{L_{1}}$       &   [N + 1,   32/64,  32/64,  $N (N(N-1)/2 + 1)$]    &  SiLU         \\
 $\alpha(t)$       &   [1,   16,   1 ]    &  SiLU         \\
  \hline
\end{tabular}
\end{table}\\
The layer width depends on the dimension $N$ and is set to 32 for $N > 10$ and to 64 otherwise.  Here,  SiLU refers to the Sigmoid Linear Unit activation function. The Plain Flow Model is implemented as a multilayer perceptron (MLP),  where the number of layers and their widths are functions of the input dimension $N$.  Initially,  the network architecture follows the layer dimensions $[N + 1, 128, 128, 128, N]$,  with SiLU used as the activation function throughout.  The Higher Gauge Flow Models are generally trained within the framework of Riemannian Flow Matching \cite{RFM}.  However,  since the base manifold is $\mathbb{R}^{N}$, training can be performed using standard Flow Matching \cite{FM}.
\subsection{Dataset}
The datasets used in the experiment are generated Gaussian-Mixture Model (GMM) datasets with:
\begin{align*}
 &N \geq 3: n_{\text{train}} = 15{,}000,\qquad
  n_{\text{test}}  = 5{,}000\\
 &N \geq 8: n_{\text{train}} = 20{,}000,\qquad
  n_{\text{test}}  = 7{,}500\\
 &N \geq 12: n_{\text{train}} = 27{,}000,\qquad
  n_{\text{test}}  = 10{,}000
\end{align*}
i.i.d.\ samples from a mixture of \(K=3000(N \geq 3),   5000(N \geq 8),  7000(N \geq 12)\) equally-weighted Gaussian components in ambient dimension \(N\in\{3,\dots,32\}\).
The GMM datasets have the following specifications:
\begin{itemize}
  \item \textbf{Mixture weights:} \(\pi_k = \frac1K,\; k=0,\dots,K-1\).
  \item \textbf{Covariances:} \(\Sigma_k = 0.5\,I_N\) (isotropic).
  \item \textbf{Spread parameter:} \(\beta = 25\).
  \item \textbf{Component means:} $\mu_k$ for each component index \(k\)
        \begin{enumerate}
          \item \emph{Primary axis:}
                \(a_1 = k \bmod N,\;
                  \mu_{k,a_1} = (-1)^k \,\beta.\)
          \item \emph{Secondary axis:}
                \(a_2 = (k + \lfloor K/2\rfloor) \bmod N.\)
                If \(a_2\neq a_1\) set
                \(\mu_{k,a_2} = (-1)^{k+1}\,\tfrac12\beta.\)
          \item \emph{Extra offset} (only when \(K>N\) and \(k\ge N\)): \\
                let \(b=(a_1+\lfloor k/N\rfloor)\bmod N\) and add
                \[
                  \mu_{k,b} \;{+}= \;
                  s_k\,0.1\,\beta \,\lfloor k/N\rfloor,\qquad
                  s_k=\begin{cases}
                    +1,& k\bmod3=0,\\
                    -1,& \text{otherwise}.
                  \end{cases}
                \]
        \end{enumerate}
\end{itemize}

\noindent
\textbf{Sampling:}\; Draw \(k\sim\mathrm{Cat}(\pi)\) and
\(x\sim\mathcal N\!\bigl(\mu_k,\,0.5\,I_N\bigr).\)

\subsection{Results}
The performance of the models was evaluated using training and testing loss, alongside the number of parameters.  Given the significant variations in loss across different values of $N$, the loss data was normalized relative to the Higher Gauge Flow Models.

\subsubsection{Train Loss}
The Higher Gauge Flow Model consistently outperformed both the Gauge Flow Model and the Plain Model across all dimensions $N$. Notably,  the magnitude of the performance difference diminished as the dimension $N$ increased.
\hspace*{-2.5cm}
\begin{figure}[h!]
  \centering
  \includegraphics[width=0.6\textwidth]{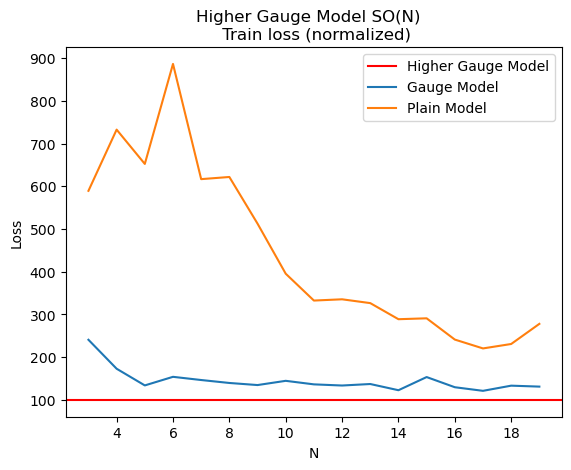}
  \caption{Train Loss Comparison (lower is better): The train loss was normalized to the loss of the Higher Gauge Model.  Loss values are shown for several dimensions $N$.}
  \label{fig:train_loss}
\end{figure}
\newpage

\subsubsection{Test Loss}
Similar to the training loss,  the testing loss consistently demonstrated the superior performance of the Higher Gauge Flow Models compared to the Gauge Flow Model and the Plain Flow Model.  A similar trend – a decreasing difference in performance – was observed with increasing dimension $N$. 

\begin{figure}[h!]
  \centering
  \includegraphics[width=0.6\textwidth]{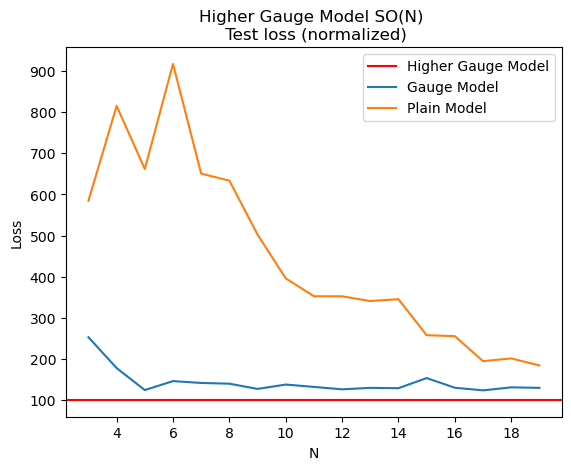}
  \caption{Test Loss Comparison (lower is better): The test loss was normalized to the loss of the Higher Gauge Model.  The test loss values are shown for several dimensions $N$.}
  \label{fig:test_loss}
\end{figure}
\newpage

\subsubsection{Number of Parameters}
\begin{figure}[h!]
  \centering
  \includegraphics[width=0.6\textwidth]{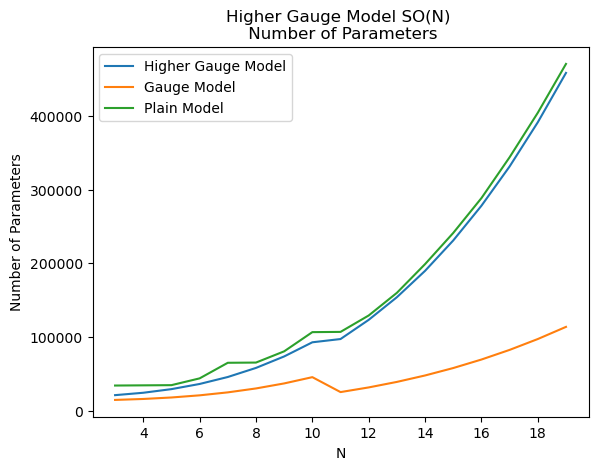}
  \caption{Number of Parameters}
  \label{fig:NumParams}
\end{figure}

As shown in Figure~\ref{fig:NumParams}, the Plain Flow Model utilized a slightly higher number of parameters than the Higher Gauge Flow Models across all dimensions $N$.  The Gauge Flow Model,  conversely,  employed significantly fewer parameters,  particularly at higher values of $N$.

\section{Related Work}
As of the author's current knowledge,  no existing study has yet connected the theory of $L_\infty$-algebras directly to methods in deep learning,  nor has Higher Gauge Theory been explicitly linked to contemporary neural-network techniques.  
By contrast, there is a rapidly growing body of work that applies ideas from Category Theory -the mathematical foundation underlying Higher Gauge Theory- to machine learning.  This line of research,  often grouped under \emph{Categorical Deep Learning},  offers a unified formalism for standard deep-learning primitives and algorithms \cite{FundamentalComponentsDeepLearning,positioncategoricaldeeplearning,CategoricalFoundations,CategoricalFoundationsDeepLearning,CatgoryTheoryMachineLearning}.  
Within this framework, key procedures such as back-propagation and gradient descent can be phrased categorically \cite{BackpropFunctor}, and constructions like \emph{lenses} provide elegant representations of supervised-learning pipelines \cite{LensesLearners}.\\

\noindent The interface between Higher Category Theory and deep learning is still largely unexplored.  Notable exceptions include the homotopy-theoretic treatment of neural networks in \cite{HomotopyTheoryNeuralNetworks} and the higher category theoretic viewpoint on network composition developed in \cite{OperadsDeepLearning}.  These works suggest that tools from higher algebra may offer principled ways to reason about network architectures and their compositional properties.

\section{Outlook}
Several promising directions deserve further investigation.  A central challenge is to incorporate higher group symmetries-those naturally associated with $L_\infty$-algebras -into neural architectures.  In general, integrating an arbitrary $L_\infty$-algebra is difficult,  and concrete higher groups are often unavailable \cite{IntegratingLInfityAlgebras}.  
A tractable starting point is the well-studied \emph{string 2-group},  arising from the string 2-Lie algebra \cite{String2LieAlgebra},  which can be viewed as a strict 2-term $L_\infty$-algebra.  Another practical avenue is to work with crossed modules of Lie algebras \cite{CrossedModules,LieAlgebraCrossedModules}, which likewise realise strict $L_\infty$-structures.

\noindent A complementary strategy is to generalise Higher Gauge Flow Models to the setting of $E\!\,L_\infty$-algebras \cite{ELInfinityAlgebras}.  Such a generalisation could furnish a principled way to encode higher-symmetry constraints directly in the data or the model,  potentially leading to architectures that respect rich algebraic structures inherent in scientific and geometric data.

\phantom{\cite{MathematicalGaugeTheory, DifferentialGeometry, DifferentialGeometryManifoldConnections,DifferentialGeometryManifold,SmoothManifolds,Manifolds,GeometryTopologyPhysics,FibreBundle,LieGroupsLieAlgebras,LieAlgebras, CNFs, GNN,GCNNs,GECNNs}}

\printbibliography

\end{document}